\documentclass{article}
\usepackage{acl}

\usepackage[utf8]{inputenc} 
\usepackage[T1]{fontenc}    

\usepackage{times,latexsym}
\usepackage{url}
\usepackage{graphicx}
\usepackage{booktabs}
\usepackage{diagbox} 
\usepackage{xcolor}
\usepackage{amsmath}
\usepackage{xspace}
\usepackage{multirow}
\usepackage{tabularx}
\usepackage{subcaption}
\usepackage{colortbl}

\usepackage{pgfplots}
\pgfplotsset{compat = newest}
\usepgfplotslibrary{patchplots,colormaps,groupplots}

\usepackage{cleveref}

\usepackage{amssymb}
\usepackage{pifont}
\newcommand{\cmark}{\ding{51}\xspace}%
\newcommand{\xmark}{\ding{55}\xspace}%

\newcommand{\Edge}{\textcolor{red}{\textit{Edge}}\xspace}%
\newcommand{\Ideal}{\textcolor{teal}{\textit{Ideal}}\xspace}%
\newcommand{\Confounder}{\textcolor{olive}{\textit{Confounder}}\xspace}%
\newcommand{\edgebox}[1]{\cellcolor{red!#1}}
\newcommand{\idealbox}[1]{\cellcolor{teal!#1}}
\newcommand{\confounderbox}[1]{\cellcolor{olive!#1}}

\newcommand{\method}[0]{Guideline-Centered Annotation Methodology\xspace}
\newcommand{\methodabbrv}[0]{GCAM\xspace}
\newcommand{\stdabbrv}[0]{SAM\xspace}

\title{Let Guidelines Guide You: \\ A Prescriptive Guideline-Centered Data Annotation Methodology}

\author{Federico Ruggeri$^1$,
  Eleonora Misino$^1$,
  Arianna Muti$^2$,
  Katerina Korre$^2$ \\
  \bf
  Paolo Torroni$^1$,
  Alberto Barrón-Cedeño$^2$ \\
  $^1$DISI, University of Bologna \\
  $^2$DIT, University of Bologna \\
  $^1$\texttt{\{federico.ruggeri6, eleonora.misino2, p.torroni\}@unibo.it} \\
  $^2$\texttt{\{arianna.muti2, aikaterini.korre2, a.barron\}@unibo.it} \\
  }

\begin{document}

\maketitle

\begin{abstract}

We introduce the \method (\methodabbrv), a novel data annotation methodology designed to report the annotation guidelines associated with each data sample. 
Our approach addresses three key limitations of the standard prescriptive annotation methodology by reducing the information loss during annotation and ensuring adherence to guidelines.
Furthermore, \methodabbrv enables the efficient reuse of annotated data across multiple tasks. 
We evaluate \methodabbrv in two ways: (i) through a human annotation study and 
(ii) an experimental evaluation with several machine learning models. 
Our results highlight the advantages of \methodabbrv from multiple perspectives, demonstrating its potential to improve annotation quality and error analysis.

\end{abstract}

\noindent \textbf{Warning}: This paper contains examples of offensive and explicit content.

\section{Introduction}
\label{sec:intro}

The widespread application of machine-learning models, and, in particular, large language models (LLMs), led to a demand surge in data availability, which, consequently, also dictated a rapid focus shift towards data quality and regulations.
Several contributions underscored the importance of clear and well-structured guidelines for producing high-quality data annotations, including annotators eligibility~\cite{zhang-etal-2023-needle}, disagreement resolution~\cite{davani-etal-2022-dealing,sandri-etal-2023-dont}, human evaluation~\cite{krishna-etal-2023-longeval}, checklists for data maintainability and reproducibility~\cite{gebru-et-al-2021-datasheets,pushkarna-et-al-2022-datacards,diaz-et-al-2022-crowdworksheets}, recommendations for mitigating harmful biases or information leakage~\cite{klie-etal-2024-dataset-annotation-quality}, and data annotation methodologies~\cite{paullada-et-al-2021-data-discontent,peng-et-al-2021-mitigating,santy-etal-2023-nlpositionality}.
Researchers have recently framed data annotation methodologies into distinct annotation paradigms. 
Each paradigm is designed to support a specific intended use of the collected data. 
These paradigms provide recommendations on how to structure the annotation process to address annotation issues, a well-known challenge in subjective tasks like hate speech detection~\cite{plank-etal-2014-linguistically, sap-etal-2022-annotators} and subjectivity detection~\cite{riloff-wiebe-2003-learning,savinova-moscoso-del-prado-2023-analyzing}.
\citet{rottger-etal-2022-two} were the first to propose the prescriptive and descriptive paradigms.
The prescriptive annotation paradigm enforces a shared belief of the task conveyed into a unique set of annotation guidelines on which annotators are instructed.
In contrast, the descriptive paradigm favors annotators' perspectives and beliefs, which can be reflected in the task.
Another notable annotation paradigm is the perspectivist one~\cite{cabitza-et-al-2023-perspectivist-turn}, where multiple annotation ground truths for the same data sample are allowed while sharing a unique set of annotation guidelines similar to the prescriptive paradigm.

\begin{table*}[!tb]
    \centering
    \footnotesize
    \begingroup
    \renewcommand{\arraystretch}{1.1} 
    \begin{tabular}{lll}
    \toprule
       \textbf{Term}  &  \textbf{Symbol} & \textbf{Definition} \\
       \midrule
        Data sample & $x$ & Textual sample to be annotated. \\
        Guideline & $g$ & Textual definition describing an annotation criterion. \\
        Guideline set (or guidelines) & $\mathcal{G}$ & Set of guidelines. \\
        Guideline subset & $\mathcal{G}_x$ & Guideline set associated with sample $x$, $\mathcal{G}_x \subseteq \mathcal{G}$. \\
        Class set & $\mathcal{C}$   &  Set of classes associated with a classification task. \\
        Class subset & $\mathcal{C}_x$ & Class set associated with sample $x$, $\mathcal{C}_x \subseteq \mathcal{C}$. \\ 
        \stdabbrv annotation function & $f$ & Function mapping a data sample $x$ to a class subset $\mathcal{C}_x$. \\
        \methodabbrv annotation function & $h$ & Function mapping a data sample $x$ to a guideline subset $\mathcal{G}_x$. \\
        \methodabbrv class grounding function & $r$ & Function mapping a guideline $g \in \mathcal{G}_x$ to a class $c \in \mathcal{C}$. \\
        \bottomrule
    \end{tabular}
    \endgroup
    \caption{Glossary of the main concepts discussed in this article.}
    \label{tab:glossary}
\end{table*}

All the existing annotation methodologies following these paradigms are centered on annotating data samples with class labels~\cite{sap-etal-2022-annotators,aldayel-etal-2021-stance-detection,kenyon-dean-etal-2018-sentiment,antici-etal-2024-corpus-sentence}.
We denote this approach as the Standard Annotation Methodology (\stdabbrv).
When considering a paradigm that relies on annotation guidelines, like the prescriptive and the perspectives ones, we argue that \stdabbrv has three main limitations that can notably affect the quality of collected data: (i) the \textit{information loss} during data annotation, where the specific guidelines used by annotators are not explicitly recorded; (ii) the \textit{lack of transparency} in evaluating annotator adherence to the guidelines, making it difficult to assess whether annotation decisions were based on the provided instructions or subjective interpretations; (iii) the \textit{task-specific annotations} resulting from \stdabbrv, which ties annotations to a task-specific class set, limiting the reuse of annotated data for other tasks unless undergoing additional human annotation efforts;
and (iv) \textit{Model Adherence to Guidelines}, where models are trained with class supervision only despite using data annotated according to specific guidelines, thus making unclear to what extent model learning aligns with the task description conveyed by guidelines.
We elaborate on our hypothesis by analyzing the prescriptive annotation paradigm, where convergence to the shared belief is an indicator of data quality and annotation guidelines are a mandatory requirement for annotation~\cite{rottger-etal-2022-two}.

In this work, we formalize and discuss the three main limitations of \stdabbrv  (\S\ref{sec:background}) and propose a novel annotation methodology, named \method (\methodabbrv), to address them (\S\ref{sec:method}).
Unlike \stdabbrv, \methodabbrv decouples the data annotation from the class set by requiring annotators to map data samples to relevant parts of guidelines rather than directly assigning class labels. 
Class labels are then associated with selected guideline parts based on the given problem formulation, which researchers define.
This two-stage process provides fine-grained guideline information for each annotation, allowing for transparency in evaluating annotation quality and flexibility in reusing annotated data across multiple tasks.
We evaluate \methodabbrv through a human annotation study (\S\ref{sec:exp:human_ann}) and machine learning experiments (\S\ref{sec:exp:model_align}), comparing it to \stdabbrv across several dimensions. 
Our results showcase that \methodabbrv improves the transparency of annotated datasets and enhances the consistency between the data annotation and model training.
We release our data, code and annotation results in a repository.\footnote{
\url{https://github.com/federicoruggeri/gcam}.
}

\section{Background and Motivation} \label{sec:background}

\begin{figure*}[!t]
     \centering
     \includegraphics[width=2.0\columnwidth]{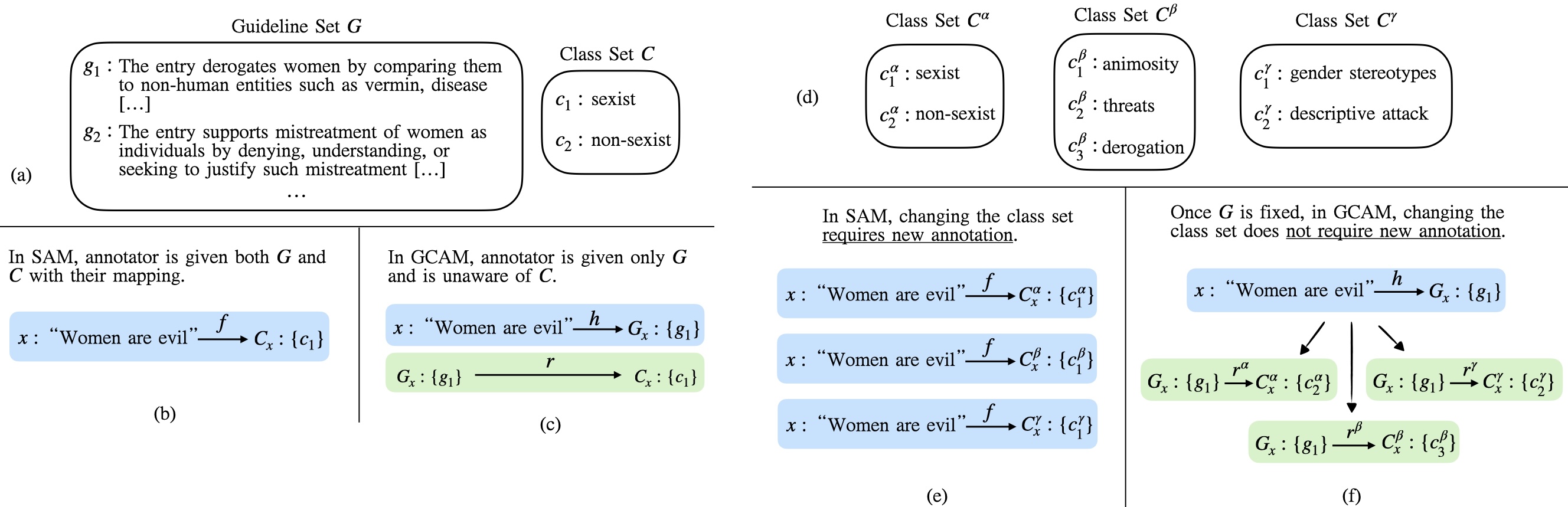}
\caption{\textbf{(a)} Example of guideline set $\mathcal{G}$ and class set $\mathcal{C}$ for hate speech detection taken from~\citet{kirk-et-al-2023-edos}.. \textbf{(b)} In \stdabbrv, the annotator knows the mapping between $\mathcal{G}$ and $\mathcal{C}$ and maps the data sample $x$ with the class subset $\mathcal{G}_x$ (blue box). \textbf{(c)} In \methodabbrv, the annotator only knows $\mathcal{G}$ and maps $x$ to $\mathcal{G}_x$ (blue box). Then, $x$ is mapped to $\mathcal{C}_x$ via the class grounding function $r$ relating $\mathcal{G}_x$ and $\mathcal{C}_x$ (green box). 
\textbf{(d)} Different class sets for the same $\mathcal{G}$. 
\textbf{(e)} In \stdabbrv, changing the class set requires new annotation; while 
\textbf{(f)} 
\methodabbrv allows annotating with different class sets via their corresponding class grounding functions (green boxes) at the cost of a single human annotation stage (blue box).
}
\label{fig:comparison_full}
\end{figure*}

We formalize \stdabbrv to highlight its properties and limitations compared to \methodabbrv.
\Cref{tab:glossary} reports the notation used in our work to avoid potential ambiguity.
According to the prescriptive paradigm~\cite{rottger-etal-2022-two}, in \stdabbrv, a group of annotators labels data samples by following a set of detailed guidelines encoding the task. 
Standard practice typically dictates a preliminary stage both for annotators and guidelines: annotators should undergo a test to be assessed for eligibility~\cite{klie-etal-2024-dataset-annotation-quality}, and guidelines should be refined iteratively through one or multiple discussion phases on identified edge cases~\cite{caselli-etal-2020-feel, guzman-monteza-2023-annotation-method, jikeli-etal-2023-antisemitic-messages,kirk-et-al-2023-edos}.

In \stdabbrv, the labeling consists of mapping each data sample to one or more labels of the provided class set $\mathcal{C}$.
The class set $\mathcal{C}$ is task-specific and may vary according to the chosen problem formulation. 
Formally, given a dataset $\mathcal{X} = \{x_1, x_2, \dots, x_{\mid \mathcal{X} \mid}  \}$ and a class set $\mathcal{C} = \{c_1, c_2, \dots, c_{\mid \mathcal{C} \mid} \}$, annotators map a data sample $x \in \mathcal{X}$ to a specific class subset $\mathcal{C}_x \subseteq \mathcal{C}$ by relying on the guideline set $\mathcal{G}= \{g_1, g_2, \dots, g_M \}.$\footnote{In a multi-class classification setting, the sample-specific class subset contains only one class (i.e., $\mathcal{C}_x = \{ c_x \}$).} 
The annotators are provided with the mapping between $\mathcal{G}$ and $\mathcal{C}$ to ground their annotations.
The annotation can be seen as a function that maps a data sample $x$ to its class subset $\mathcal{C}_x$:
\begin{equation} \label{eq:standard_}
    f: \mathcal{X} \rightarrow \mathcal{P}(\mathcal{C}),
\end{equation}

with $\mathcal{P}(\mathcal{C})$ the power set of the class set.
An example of \stdabbrv is depicted in \Cref{fig:comparison_full}(a,b), where the annotator implicitly uses the mapping between $\mathcal{G}$ and $\mathcal{C}$ to label $x$ as \texttt{sexist}.

\paragraph{Assumptions.}
For both \stdabbrv and \methodabbrv we assume that i)
the guideline set $\mathcal{G}$ is fixed throughout data annotation according to the prescriptive paradigm~\cite{rottger-etal-2022-two}, and ii)
data annotation guidelines can be framed as a set of criteria $g$.
Criteria are used to reduce annotators' effort and provide systematic instructions that are easy to assimilate and follow~\cite{sabou-etal-2014-corpus, ruggeri-etal-2023-prescriptive-annotation}.

\subsection{Limitations} \label{sec:sam-limitations}

In \stdabbrv, annotators do not report the guideline subset they rely upon for annotating each data sample.
We see this information loss as the cause of the following issues undermining annotation quality.

\paragraph{Lack of Transparency}
Evaluating annotation quality is crucial in ensuring the reliability of annotated data~\cite{klie-etal-2024-dataset-annotation-quality}.
The annotation quality can be influenced by three main factors: i) intrinsic complexity of the task, ii) guidelines clarity and coverage of the task, and iii) annotators' understanding of the guidelines.
While task complexity is an inherent property of a problem, the other two factors can be addressed to enhance annotation quality.
However, in \stdabbrv, verifying annotators' adherence to the guidelines with only class label annotations is challenging. 
This has implications at two distinct stages.
First, during the annotator eligibility assessment, it is difficult to determine whether they have assigned their labels following the provided guidelines or using their subjective criteria.
Second, during annotation, it is unclear whether errors stem from unclear guidelines or misinterpretation by the annotator. 

\paragraph{Task-specific Annotations}
\stdabbrv's reliance on performing data annotation using a task-specific class set limits the reuse of annotated data across tasks. 
In particular, when a different task shares the same guidelines but employs a different class set, the previously annotated data cannot be reused and another data annotation stage is mandatory.

\paragraph{Model Adherence to Guidelines}
Machine Learning (ML) tasks are typically structured to enable a model to learn the relation between input features ($x$) and target variables ($\mathcal{C}_x$).  
When training a model with annotated data, the underlying assumption is that the optimal mapping learned by the model from $x$ to $\mathcal{C}_x$ is the one adherent to the guideline set. 
However, no explicit constraints prevent the model from learning spurious correlations from data and, more in general, exploiting shortcuts~\cite{geirhos-et-al-2020-shortcut-learning}.
In principle, some models could potentially leverage the knowledge encoded in the guidelines subset $\mathcal{G}_x$ similarly to how external knowledge is integrated into machine learning models~\cite{von-rueden-etal-2023-informed-ml}. 
However, they need a dataset that contains this information to explicitly assess the knowledge integration process, which cannot be created with \stdabbrv.

\section{\methodabbrv} \label{sec:method}
We describe \method (\methodabbrv), a novel methodology aimed at enriching the annotation by keeping track of the guidelines associated with data samples. 
Conversely to \stdabbrv, in \methodabbrv, annotators do not have access to the class set $\mathcal{C}$, but only to the guidelines $\mathcal{G}$.
This ensures the annotation to be \textit{adherent to $\mathcal{G}$} and  \textit{independent from $\mathcal{C}$ by definition}.

\subsection{Formalization} \label{sec:gc:formalization}
\methodabbrv comprises two stages. 
First, annotators are provided with guideline $\mathcal{G}$ and data $\mathcal{X}$, and are tasked with mapping each data point to their guideline subset $\mathcal{G}_x$. 
In the second phase, which does not involve annotators, one of the possible mappings between $\mathcal{G}$ and $\mathcal{C}$ is applied to associate samples to class labels.
Formally, \methodabbrv can be seen as the composition of two functions:
\begin{align}
    h &: \mathcal{X} \rightarrow \mathcal{P}(\mathcal{G}),
    \label{eq:two-stage_formulation_h}\\
    r &: \mathcal{G} \rightarrow \mathcal{C}.
    \label{eq:two-stage_formulation_r}
\end{align}
\Cref{eq:two-stage_formulation_h} represents the annotator's function that maps a data sample $x$ to a guideline subset $\mathcal{G}_x$, while \Cref{eq:two-stage_formulation_r} is the \textit{class grounding function} mapping each guideline $g$ to the corresponding class.
The function $h$ is a black box proper of each annotator, similar to $f$ in \Cref{eq:standard_}, while the function $r$ is known by design.
\Cref{fig:comparison_full}(c) illustrates \methodabbrv.
In the first stage, the human annotator maps $x = \textit{"Women are evil"}$ to $\mathcal{G}_x = \{ g_1 \}$.
Then, in the second stage, $\mathcal{G}_x$ is mapped to $\mathcal{C}_x = \{ \texttt{sexist} \}$ via $r$.

\subsection{Properties} \label{sec:gc:properties}
We describe three major properties of \methodabbrv addressing the issues of \stdabbrv (\S\ref{sec:sam-limitations}).

\paragraph{Transparent Adherence to Guidelines}
A key aspect of \methodabbrv is that annotators are unaware of the class set $\mathcal{C}$ and the class grounding function $r$.
This enforces adherence by definition to the guidelines during the annotation process, as the annotators do not have access to other information on the task.
In practice, \methodabbrv requires guidelines to not contain any reference to $\mathcal{C}$ and $r$.
Another essential part of the annotation procedure is the discussion phase, as it aims to reach an agreement on edge cases.
In \methodabbrv, discussions revolve exclusively around the guideline subset $\mathcal{G}_x$ and do not involve any reasoning on $\mathcal{C}$, as annotators lack access to it during the annotation phase. 
In this setting, the convergence of annotators' points of view can be effectively evaluated by inspecting $\mathcal{G}_x$.

\paragraph{Different Task Annotations}
\methodabbrv enables using the annotated dataset for different task annotations that share the same guidelines. 
These annotations may stem from a taxonomy \cite{kirk-et-al-2023-edos} or account for discrepancies in how a task should be addressed, as it is often the case in subjective tasks~\cite{antici-etal-2024-corpus-sentence}.
Consider an example with three different class sets $ \mathcal{C}^\alpha, \mathcal{C}^\beta, \mathcal{C}^\gamma$  corresponding to three possible formulations of the hate speech detection task (Figure~\ref{fig:comparison_full}(d)). 
In GCAM (Figure~\ref{fig:comparison_full}(f)) a human annotator annotates the data sample $x$ with corresponding guiline subset $\mathcal{G}_x$. 
The resulting $\mathcal{G}_x$ is mapped to the three class subsets $\mathcal{C}_x^\alpha$, $\mathcal{C}_x^\beta$, and $\mathcal{C}_x^\gamma$ via $r$ by only paying the cost of a single human annotation stage.
This is a key advantage of \methodabbrv over \stdabbrv, in which changing the class set would require new annotation stages (Figure~\ref{fig:comparison_full}(e)). 

\paragraph{Model Adherence to Guideline}
Datasets created following \methodabbrv report explicitly the knowledge encoded in the guidelines $\mathcal{G}$.
This allows ML models to be effectively trained to leverage this information. 
We highlight three main advantages of training a model on \methodabbrv datasets.
First, we effectively prevent the model from learning spurious correlations between $x$ and the class subset $\mathcal{C}_x$, as the model has no access to $\mathcal{C}_x$ during training.
Second, the decoupling between $\mathcal{G}$ and the class set $\mathcal{C}$ 
allows for using the model for different task formulations, as its predictions can be mapped to different class sets relying on the corresponding class grounding functions.
Third, we can evaluate the model's performance by assessing its alignment with $\mathcal{G}$ by analyzing disparities between predicted and ground-truth $\mathcal{G}_x$, fostering a deeper comprehension of the model.


\section{Related Work} \label{sec:related}

Annotation paradigms have been recently proposed for data annotation to encode one shared view, such as the prescriptive paradigm~\cite{rottger-etal-2022-two}, or different beliefs about a phenomenon, such as the descriptive~\cite{rottger-etal-2022-two}, and perspectivist~\cite{cabitza-et-al-2023-perspectivist-turn} paradigms.
The goal of annotation paradigms is to regulate data annotation for a specific intended use of collected data.
In the prescriptive paradigm, this objective is achieved using annotation guidelines~\cite{benikova-etal-2014-nosta, sabou-etal-2014-corpus, zaghouani-etal-2014-large, krishna-etal-2023-longeval}. 
Notable examples address subjective tasks, such as hate speech~\cite{salminen-etal-2019-online-hate,sap-etal-2022-annotators,davani-etal-2023-hate}, stance detection~\cite{luo-etal-2020-detecting,aldayel-etal-2021-stance-detection}, sentiment analysis~\cite{waseem-hovy-2016-hateful,kenyon-dean-etal-2018-sentiment}, and subjectivity detection~\cite{antici-etal-2024-corpus-sentence}.
In all these approaches, annotators do not keep track of the annotation guidelines they rely on to annotate data samples.
Therefore, although of crucial importance, the mapping between guidelines and data samples is lost during the annotation.

To the best of our knowledge, \citet{jikeli-etal-2023-antisemitic-messages} is the only work that designs a data annotation methodology that enforces explicit grounding to annotation guidelines.
In particular, annotators must specify which guideline applies to each data sample regarding antisemitism detection.  
Our work differs from~\citeauthor{jikeli-etal-2023-antisemitic-messages} in several aspects.
First, the work of \citeauthor{jikeli-etal-2023-antisemitic-messages} is limited to a specific issue within social media content analysis and does not present a general formulation of the annotation process.
Second, the proposed data annotation methodology relies on reporting both  $\mathcal{G}_x$ and $\mathcal{C}_x$ for each data sample $x$.
This process suffers from an annotation shortcut where annotators may first define $\mathcal{C}_x$ and subsequently $\mathcal{G}_x$, thus reversing the intended annotation order.
Furthermore, the annotation methodology is tight to a specific class set $\mathcal{C}$ and has increased annotation workload due to reporting both $\mathcal{G}_x$ and $\mathcal{C}_x$.
We elaborate more on these two aspects in~\S\ref{sec:discussion}.



\section{\methodabbrv vs. \stdabbrv in Annotation}
\label{sec:exp:human_ann}

We design a human annotation study to empirically compare \methodabbrv with \stdabbrv.
In particular, we consider the challenging task of subjectivity detection~\cite{riloff-wiebe-2003-learning}.

\paragraph{Setup.}
We follow the annotation methodology of~\citet{antici-etal-2024-corpus-sentence} and instruct four annotators for sentence-level subjectivity detection.
In particular, annotators label a sentence as subjective or objective ($\mathcal{C} = \{\texttt{SUBJ}, \texttt{OBJ}\}$) using a guideline set of 12 guidelines (i.e., $\vert \mathcal{G} \vert = 12$).
We split the four annotators into two pairs: $p_{\stdabbrv}$ and $p_{\methodabbrv}$.
$p_{\stdabbrv}$ annotates following \stdabbrv while $p_{\methodabbrv}$ annotates using \methodabbrv.
We use the original guidelines of~\citet{antici-etal-2024-corpus-sentence} for \stdabbrv\footnote{$\{g_1, \dots, g_5$\} for \texttt{SUBJ} and $\{g_6, \dots, g_{12}$\} for \texttt{OBJ}.}, while we build a variant of these guidelines for \methodabbrv to exclude information about $\mathcal{C}$ and $r$.
We collect 200 training samples from NewsSD-ENG~\cite{antici-etal-2024-corpus-sentence} and split them into two data batches, $b_1$ and $b_2$, each with 100 samples, to limit data and annotator selection bias and ensure a sound evaluation setting.
Each annotator pair labels all data batches, totaling 200 samples per pair.
We compare \methodabbrv and \stdabbrv both quantitatively by computing inter-annotator agreement (IAA) and qualitatively by reporting annotator feedback and analyzing annotator discussion.

\paragraph{Quantitative Results.}
We compute the IAA measured as Krippendorff's alpha~\cite{krippendorff2004measuring} for each annotator pair.
Regarding $\mathcal{C}_x$, the IAA in $b_1$ is $54.00$ for \stdabbrv and $46.14$ for \methodabbrv, while, in $b_2$, is $62.00$ for \stdabbrv and $57.70$ for \methodabbrv.
These observations show that IAA is comparable for the same pair across batches and different pairs on the same batch and that annotating with \methodabbrv is comparable to \stdabbrv.\footnote{We refer to Krippendorff's alpha intervals.}
Regarding $\mathcal{G}_x$, the IAA is $28.89$ in $b_1$ and $38.86$ in $b_2$, highlighting that agreeing on a wide set of guidelines is notably harder.
Moreover, we evaluate annotation effort by measuring annotation time per batch.
On average, $p_{\stdabbrv}$ takes $\sim$60min to annotate a batch, while $p_{\methodabbrv}$ takes $\sim$150min.
As expected, \methodabbrv has time overhead for annotation. 
This is due to the requirement for annotators to report $\mathcal{G}_x$, which is usually far more fine-grained than $\mathcal{C}_x$ and, thus, more time-consuming to assimilate.
The annotation effort in \methodabbrv is required to enforce a transparent annotation that allows evaluating annotators' convergence to $\mathcal{G}$.
We elaborate more in the following paragraph.

\paragraph{Qualitative Results.}

 




We collect annotators' feedback during each batch to evaluate their adherence to the annotation task. 
Interestingly, $p_{\stdabbrv}$ annotators report relying more on their subjective criteria and less on $\mathcal{G}$ throughout the study. 
Specifically, annotators consult $\mathcal{G}$ mainly for challenging data samples, while they often address easier ones without referring to $\mathcal{G}$. 
This does not necessarily indicate a deviation from $\mathcal{G}$, as their internalized knowledge from training may still align with $\mathcal{G}$.
Nonetheless, in \stdabbrv,  we cannot further analyze annotators' adherence to $\mathcal{G}$.
In contrast, $p_{\methodabbrv}$ annotators always rely on $\mathcal{G}$ by design, allowing for direct inspection of the selected $\mathcal{G}_x$.
These observations raise a major point of discussion: \textit{to what extent does a good IAA on $\mathcal{C}$ reflect a correct application of the annotation paradigm?}
We argue that it is hardly ever satisfactory enough.
In particular, we analyze disagreement in \methodabbrv concerning $\mathcal{G}_x$ to demonstrate that (i) the majority of cases still agree on $\mathcal{C}$ and (ii) that there is more than one reason annotators disagree, highlighting different types of edge cases.\footnote{We apply majority voting on $\mathcal{C}_x$ to obtain a single prediction.}
$66$ out of $115$ disagreement cases ($\sim$57\%) agree on $\mathcal{C}_x$ while reporting disjoint $\mathcal{G}_x$ sets ($58$) or where one annotator's $\mathcal{G}_x$ subsumes the other's ($8$).
In the remaining $49$ cases, $47$ are due to disjoint $\mathcal{G}_x$ sets and $2$ concern subsumption.
After discussion, $72$ cases are resolved by selecting one annotator's $\mathcal{G}_x$, $15$ by taking the union of the sets, and $25$ by agreeing on a different $\mathcal{G}_x$.
These results also underscore the difficulty of achieving consensus on a shared belief, as the prescriptive paradigm dictates, even when a transparent annotation like \methodabbrv is enforced.
We remark that these considerations cannot be addressed via \stdabbrv, but only when providing information about $\mathcal{G}_x$ as \methodabbrv provides.
This is a major advantage of \methodabbrv on \stdabbrv when analyzing data quality.

\section{Model Alignment with Guidelines}
\label{sec:exp:model_align}


\begin{table*}[!tb]
\centering
\scriptsize
\bgroup
\def\arraystretch{1.2}%
\resizebox{\linewidth}{!}{
\begin{tabular}{ll|rrr||rrrrrr}
\toprule
{} & {} & \multicolumn{1}{c}{\stdabbrv} & \multicolumn{2}{c||}{\methodabbrv} & \multicolumn{3}{c}{\stdabbrv} &\multicolumn{3}{c}{\methodabbrv}\\
\noalign{\vskip -4pt} & & & & & & \\

{} & {} & \multicolumn{1}{c}{\textit{Binary}} & \multicolumn{1}{c}{\textit{Fine-grained}} & \multicolumn{1}{c||}{\textit{Text-based}${}^\alpha$} & \multicolumn{1}{c}{\textit{Mistral}} & \multicolumn{1}{c}{\textit{Llama}} & \multicolumn{1}{c}{\textit{Phi}} & \multicolumn{1}{c}{\textit{Mistral}} & \multicolumn{1}{c}{\textit{Llama}} & \multicolumn{1}{c}{\textit{Phi}} \\ 
\midrule
\multirow{2}{*}{ToS$_{A}$}  
  & $\mathcal{C}_x$ & $71.61_{\pm 6.76}$ & $65.31_{\pm 4.15}$ & $\mathbf{80.14_{\pm 5.29}^{*,**}}$ & $57.07$ & $62.72$ & $64.10$ & $0.59$ & $36.04$ & $37.80$ \\
  & \cellcolor[gray]{0.9}$\mathcal{G}_x$ & \cellcolor[gray]{0.9}- & \cellcolor[gray]{0.9}$18.33_{\pm 5.22}$ & \cellcolor[gray]{0.9}$\mathbf{37.49_{\pm 5.28}^{\text{ -},**}}$ & \cellcolor[gray]{0.9}- & \cellcolor[gray]{0.9}- & \cellcolor[gray]{0.9}- & \cellcolor[gray]{0.9}$0.11$ & \cellcolor[gray]{0.9}$1.44$ & \cellcolor[gray]{0.9}$3.89$ \\
\multirow{2}{*}{ToS$_{CH}$}  
  & $\mathcal{C}_x$ & $87.96_{\pm 3.19}$ & $83.61_{\pm 2.90}$ & $\mathbf{90.23_{\pm 1.84}^{ns,**}}$ & $62.43$ & $60.80$ & $62.09$ & $1.70$ & $21.31$ & $8.35$ \\
  & \cellcolor[gray]{0.9}$\mathcal{G}_x$ & \cellcolor[gray]{0.9}- & \cellcolor[gray]{0.9}$34.58_{\pm 4.61}$ & \cellcolor[gray]{0.9}$\mathbf{44.92_{\pm 9.93}^{\text{ -},**}}$ & \cellcolor[gray]{0.9}- & \cellcolor[gray]{0.9}- & \cellcolor[gray]{0.9}- & \cellcolor[gray]{0.9}$0.08$ & \cellcolor[gray]{0.9}$3.06$ & \cellcolor[gray]{0.9}$2.16$ \\
\multirow{2}{*}{ToS$_{CR}$}  
  & $\mathcal{C}_x$ & $\mathbf{79.02_{\pm 2.77}}$ & $67.85_{\pm 5.33}$ & $78.14_{\pm 3.76}^{ns,**}$ & $55.76$ & $56.64$ & $59.96$ & $2.46$ & $27.74$ & $25.23$ \\
  & \cellcolor[gray]{0.9}$\mathcal{G}_x$ & \cellcolor[gray]{0.9}- & \cellcolor[gray]{0.9}$17.37_{\pm 3.59}$ & \cellcolor[gray]{0.9}$\mathbf{27.21_{\pm 5.36}^{\text{ -},**}}$ & \cellcolor[gray]{0.9}- & \cellcolor[gray]{0.9}- & \cellcolor[gray]{0.9}- & \cellcolor[gray]{0.9}$0.46$ & \cellcolor[gray]{0.9}$1.18$ & \cellcolor[gray]{0.9}$1.17$ \\
\multirow{2}{*}{ToS$_{LTD}$}  
  & $\mathcal{C}_x$ & $75.87_{\pm 2.36}$ & $73.70_{\pm 1.79}$ & $\mathbf{80.39_{\pm 1.40}^{**,**}}$ & $60.71$ & $55.79$ & $58.00$ & $8.74$ & $21.16$ & $7.87$ \\
  & \cellcolor[gray]{0.9}$\mathcal{G}_x$ & \cellcolor[gray]{0.9}- & \cellcolor[gray]{0.9}$19.87_{\pm 2.70}$ & \cellcolor[gray]{0.9}$\mathbf{21.49_{\pm 1.41}^{\text{ -},ns}}$ & \cellcolor[gray]{0.9}- & \cellcolor[gray]{0.9}- & \cellcolor[gray]{0.9}- & \cellcolor[gray]{0.9}$0.57$ & \cellcolor[gray]{0.9}$5.03$ & \cellcolor[gray]{0.9}$4.32$ \\
\multirow{2}{*}{ToS$_{TER}$}  
  & $\mathcal{C}_x$ & $\mathbf{86.86_{\pm 1.54}}$ & $78.80_{\pm 5.56}$ & $84.57_{\pm 4.94}^{ns,*}$ & $62.70$ & $54.86$ & $57.81$ & $2.33$ & $18.21$ & $2.63$ \\
  & \cellcolor[gray]{0.9}$\mathcal{G}_x$ & \cellcolor[gray]{0.9}- & \cellcolor[gray]{0.9}$10.16_{\pm 2.91}$ & \cellcolor[gray]{0.9}$\mathbf{20.04_{\pm 3.08}^{\text{ -},**}}$ & \cellcolor[gray]{0.9}- & \cellcolor[gray]{0.9}- & \cellcolor[gray]{0.9}- & \cellcolor[gray]{0.9}$0.53$ & \cellcolor[gray]{0.9}$0.99$ & \cellcolor[gray]{0.9}$1.19$ \\
\multirow{2}{*}{EDOS}  
  & $\mathcal{C}_x$ & $\mathbf{81.74_{\pm 0.59}}$ & $71.67_{\pm 2.78}$ & $79.87_{\pm 2.24}^{*,**}$ & $56.47$ & $67.93$ & $63.47$ & $25.85$ & $38.80$ & $28.29$ \\
  & \cellcolor[gray]{0.9}$\mathcal{G}_x$ & \cellcolor[gray]{0.9}- & \cellcolor[gray]{0.9}$\mathbf{27.09_{\pm 2.69}}$ & \cellcolor[gray]{0.9}$22.97_{\pm 3.14}^{\text{ -},**}$ & \cellcolor[gray]{0.9}- & \cellcolor[gray]{0.9}- & \cellcolor[gray]{0.9}- & \cellcolor[gray]{0.9}$9.97$ & \cellcolor[gray]{0.9}$11.27$ & \cellcolor[gray]{0.9}$5.02$ \\
\bottomrule
\end{tabular}
}
\egroup
\caption{Model performance on ToS and EDOS datasets for encoder-based models (left) and LLMs (right).  Best results are in \textbf{bold}. \\
\scriptsize{$^\alpha$We report comma-separated Wilcoxon statistical significance comparing \textit{Text-based} to \textit{Binary} and \textit{Fine-grained}, respectively. (-) stands for not applicable, (ns) for no significance, $(^{*}) \le 0.05$, $(^{**}) \le 0.01$.}
}
\label{tab:experiments:classification-merged}
\end{table*}

We assess model performance on data annotated following \stdabbrv and \methodabbrv, and we discuss the differences in error analysis when models report $\mathcal{C}_x$ (in \stdabbrv) or $\mathcal{G}_x$ (in \methodabbrv).
For this purpose, we devise two experimental settings: supervised training of encoder-based transformers and zero-shot prompting of LLMs.

\paragraph{Data.}
We consider two text classification datasets that report information about $\mathcal{G}_x$ in addition to $\mathcal{C}_x$: the Online Terms of Service (ToS-100)~\cite{ruggeri-et-al-2022-tos} and the Explainable Detection of Online Sexism (EDOS)~\cite{kirk-et-al-2023-edos} datasets.
ToS-100 contains 100 documents ($\sim$20k sentences) about unfair clause detection, where a clause is labeled as either fair (negative class) or unfair (positive class).
Documents are labeled according to five categories of unfairness. 
For each category, annotators label positive samples by reporting one or multiple reasons for unfairness, denoted as legal rationales and manually curated by legal experts.
Following~\citet{ruggeri-et-al-2022-tos}, we train a binary classifier for each of the five unfairness categories and use their corresponding legal rationales  $\mathcal{G}$ (see Appendix~\ref{app:tos}).
EDOS contains $\sim$20k social media comments annotated for three tasks of incremental specification: sexism detection, sexism categorization, and fine-grained vectors of sexism.
For the latter task, the vectors of sexism correspond to the annotation guidelines shared among all tasks.
We define $\mathcal{G}$ as this set of guidelines ($\vert \mathcal{G} \vert = 11$).
Unlike ToS-100, positive data samples in EDOS can only be associated with one guideline $g$ by design.

\subsection{Encoder-based Models} \label{sec:encoder-based}
We compare state-of-the-art models trained on data annotated with \stdabbrv or \methodabbrv.
For ToS-100, we consider LegalBERT\footnote{\texttt{nlpaueb/legal-bert-base-uncased}}~\cite{chalkidis-etal-2020-legal}, as a top-performing model on legal benchmarks~\cite{chalkidis-etal-2022-lexglue}.
Likewise, we consider RoBERTaHate\footnote{\texttt{cardiffnlp/twitter-roberta-base-hate}}~\cite{barbieri-etal-2020-tweeteval} for EDOS.

\paragraph{Problem Formulation.}
In \stdabbrv, given an input data sample $x$, the model is trained to predict $\mathcal{C}_x$.
In particular, the model does not have access to $\mathcal{G}$ and only receives feedback based on $\mathcal{C}_x$ supervision.
In our experiment, we consider a binary classification task.
In ToS, it translates to classifying a clause as unfair; in EDOS, the task is to classify a text as sexist.
We denote the \stdabbrv setting as \textit{Binary}.
Regarding \methodabbrv, we consider two classification settings: \textit{Fine-grained} and \textit{Text-based}.
In the \textit{Fine-grained} setting, a model considers each guideline $g$ as a distinct class, and no textual content of $g$ is provided.
In the \textit{Text-based} setting, we follow the general architecture of~\citet{pappas-henderson-2019} to train a model on guideline subsets $\mathcal{G}_x$ and, therefore, evaluate the impact of guideline textual content.
More precisely, given an input pair $(x, g)$, the model is trained to classify if $x$ relates to $g$. 
Thus, the model is trained to perform a textual entailment task for all $g \in \mathcal{G}$ given a data sample $x$.
In the \textit{Text-based} setting, we tokenize and encode $x$ and $g$ together and feed the resulting embedding to the classification head.\footnote{We consider other variants in Appendix~\ref{app:text-based-variants}.}

\paragraph{Evaluation.}
We perform a repeated train and test evaluation setting with 10 seed runs.
We split ToS-100 at the document level in train (80), validation (10), and test (10) partitions while we use the official data partitions for EDOS.
We fine-tune models using their default hyper-parameters and apply early stopping regularization on validation loss with patience equal to 5 epochs.
We report macro F$_1$ score on both $\mathcal{C}_x$ and $\mathcal{G}_x$ to assess model performance.
In particular, for \methodabbrv, we aggregate $\mathcal{G}_x$ predictions via $r$ to obtain the corresponding $\mathcal{C}_x$ for binary classification.
See Appendix~\ref{app:hardware} for hardware and implementation details.

\paragraph{Quantitative Results.}
Table~\ref{tab:experiments:classification-merged} (left) reports classification results on both datasets.
Regarding binary classification task ($\mathcal{C}_x$), \textit{Text-based} outperforms \textit{Binary} in two settings (ToS$_{A}$ and ToS$_{LTD}$), while reports comparable performance in the remaining ones.
In contrast, \textit{Text-based} significantly outperforms \textit{Fine-grained} in all datasets.
Regarding $\mathcal{G}_x$, \textit{Text-based} outperforms \textit{Fine-grained} in all ToS settings by a notable margin while \textit{Fine-grained} achieves better performance in EDOS.
These results highlight the benefit of integrating guideline textual content during learning.
Overall, empirical observations suggest that training a model on a dataset annotated with \methodabbrv is as effective as with \stdabbrv from a performance viewpoint. 

\paragraph{Qualitative Results.}
We inspect \textit{Binary} for \stdabbrv and \textit{Text-based} for \methodabbrv as the top-performing models to assess how the two annotation methodologies differ when analyzing model predictions.\footnote{We consider the best-performing seed of each model for a fair comparison.}
We focus on the EDOS dataset for simplicity of evaluation since task formulation for \methodabbrv is cast to a multi-class classification problem.
Regarding \stdabbrv, the confusion matrix of \textit{Binary} for $\mathcal{C}_x$ provides details about classification errors (\Cref{fig:cm-encoder-based-a}).
We observe that most errors are false positives, suggesting the presence of edge cases in the dataset that might confuse classifiers.
Compared to \stdabbrv, \methodabbrv offers a more fine-grained assessment of model errors by providing three distinct evaluation tools.
First, we compute the confusion matrix on $\mathcal{C}_x$ as done in \stdabbrv (\Cref{fig:cm-encoder-based-b}).
Second, we compute the confusion matrix on $\mathcal{G}_x$ to assess confounders between guidelines $g$ (\Cref{fig:cm-encoder-based-c}).
For instance, we can observe that $g_4$ is a strong confounder for $g_6$ samples and, likewise, $g_1$ for $g_3$.
Third, we categorize model predictions by applying $r$ to link the two confusion matrices.
We denote this tool as the \textit{grounding error types matrix}.
In particular, we identify three cases: $\mathcal{C}_x$ and $\mathcal{G}_x$ are incorrect (\Edge); $\mathcal{C}_x$ and $\mathcal{G}_x$ are correct (\Ideal); $\mathcal{C}_x$ is correct but $\mathcal{G}_x$ is incorrect (\Confounder).\footnote{The case where $\mathcal{G}_x$ is correct and $\mathcal{C}_x$ is incorrect is not possible.} 
While \stdabbrv aggregates all correct predictions on $\mathcal{C}_x$, \methodabbrv distinguishes between \textit{Ideal} and \textit{Confounder}.
This is an important characteristic of \methodabbrv, which allows for considerations on which model predictions should be labeled as correct.
For instance, we argue that \textit{Confounder} predictions should be labeled as incorrect as they represent a clear violation of the prescriptive paradigm if we consider those predictions derived by a human annotator.

\subsection{LLM Prompting}
We evaluate LLMs in the same setting of~\S\ref{sec:encoder-based}.
We experiment with three open-source state-of-the-art LLMs: Mistral3\footnote{\texttt{mistralai/Mistral-7B-Instruct-v0.3}.}, Llama3\footnote{\texttt{meta-llama/Llama-3.1-8B-instruct.}}, and Phi3-mini.\footnote{\texttt{microsoft/Phi-3-mini-4k-instruct}}

\paragraph{Problem Formulation.}
In \stdabbrv, given an input data sample $x$, an LLM is prompted to predict $\mathcal{C}_x$.
We include $\mathcal{G}$ in the prompt to provide task description and perform in-context learning~\cite{dong-etal-2024-in-context-learning}.
In \methodabbrv, given an input data sample $x$, an LLM is prompted to predict $\mathcal{G}_x$. In this case, no information about $r$ nor $\mathcal{C}$ is provided in the prompt.

\paragraph{Setup.}
We prompt LLMs in a zero-shot setting to assess the impact of the two problem formulations.
We set the temperature to zero to exclude variations in the generated response since we prompt LLMs for a classification task.
In this way, a change in the answer is attributed solely to the prompt definition.
See our code repository for details about the prompts used.

\begin{figure}[!t]
\small
\centering

\begin{minipage}{0.4\linewidth}
\centering
\begin{subtable}[t]{1\textwidth}
\centering
\setlength{\arrayrulewidth}{0.3mm}
\setlength{\tabcolsep}{5pt} 
\renewcommand{\arraystretch}{1.2}
\begin{tabular}{c|c|c}
T / P  & Neg. & Pos. \\
\hline
Neg. &  2710 &  320 \\
\hline
Pos. &  199 &  771 \\
\end{tabular}
\caption{$\mathcal{C}_x$ for \stdabbrv.} \label{fig:cm-encoder-based-a}
\end{subtable}
\end{minipage}%
\hspace{2em} 
\begin{minipage}{0.4\linewidth}
\centering
\begin{subtable}[t]{1\textwidth}
\centering
\setlength{\arrayrulewidth}{0.3mm}
\setlength{\tabcolsep}{5pt} 
\renewcommand{\arraystretch}{1.2}
\begin{tabular}{c|c|c}
T / P  & Neg. & Pos. \\
\hline
Neg. & 2673 & 357 \\
\hline
Pos. &  182 &  788 \\
\end{tabular}
\caption{$\mathcal{C}_x$ for \methodabbrv.} \label{fig:cm-encoder-based-b}
\end{subtable}
\end{minipage}
\vspace{1.5em} 

\definecolor{col1}{rgb}{0.9,0.95,1}
\definecolor{col2}{rgb}{0.7,0.85,1}
\definecolor{col3}{rgb}{0.5,0.75,1}
\definecolor{col4}{rgb}{0.3,0.65,1}
\definecolor{col5}{rgb}{0.1,0.55,1}

\begin{minipage}{1.0\linewidth}
\centering
\begin{subtable}[t]{1\textwidth}
\centering
\setlength{\arrayrulewidth}{0.3mm}
\setlength{\tabcolsep}{5pt} 
\renewcommand{\arraystretch}{1.2}
\resizebox{\columnwidth}{!}{
\begin{tabular}{c|ccccccccccc}
T / P & $g_1$ & $g_2$ & $g_3$ & $g_4$ & $g_5$ & $g_6$ & $g_7$ & $g_8$ & 
$g_9$ & $g_{10}$ & $g_{11}$ \\
\hline
\textbf{$g_1$} & \cellcolor{col5}2680 & \cellcolor{col3}41 & \cellcolor{col4}159 & \cellcolor{col1}4 & \cellcolor{col1}5 & \cellcolor{col3}40 & \cellcolor{col4}98 & \cellcolor{col1}0 & \cellcolor{col1}0 & \cellcolor{col2}10 & \cellcolor{col2}9  \\
\hline
\textbf{$g_2$} & \cellcolor{col2}20 & \cellcolor{col2}22 & \cellcolor{col1}2 & \cellcolor{col1}3 & \cellcolor{col1}1 & \cellcolor{col3}24 & \cellcolor{col1}0 & \cellcolor{col1}0 & \cellcolor{col1}0 & \cellcolor{col1}0 & \cellcolor{col1}1 \\
\hline
\textbf{$g_3$} & \cellcolor{col2}34 & \cellcolor{col1}0 & \cellcolor{col4}119 & \cellcolor{col1}0 & \cellcolor{col1}1 & \cellcolor{col3}23 & \cellcolor{col3}24 & \cellcolor{col1}0 & \cellcolor{col1}0 & \cellcolor{col1}3 & \cellcolor{col1}1 \\
\hline
\textbf{$g_4$} & \cellcolor{col2}22 & \cellcolor{col1}1 & \cellcolor{col2}22 & \cellcolor{col1}7 & \cellcolor{col1}0 & \cellcolor{col5}140 & \cellcolor{col1}0 & \cellcolor{col1}0 & \cellcolor{col1}0 & \cellcolor{col1}0 & \cellcolor{col1}0 \\
\hline
\textbf{$g_5$} & \cellcolor{col2}19 & \cellcolor{col1}2 & \cellcolor{col2}18 & \cellcolor{col1}0 & \cellcolor{col1}2 & \cellcolor{col2}14 & \cellcolor{col1}2 & \cellcolor{col1}0 & \cellcolor{col1}0 & \cellcolor{col1}0 & \cellcolor{col1}0 \\
\hline
\textbf{$g_6$} & \cellcolor{col1}7 & \cellcolor{col1}2 & \cellcolor{col1}3 & \cellcolor{col1}4 & \cellcolor{col1}0 & \cellcolor{col5}166 & \cellcolor{col1}0 & \cellcolor{col1}0 & \cellcolor{col1}0 & \cellcolor{col1}0 & \cellcolor{col1}0 \\
\hline
\textbf{$g_7$} & \cellcolor{col2}30 & \cellcolor{col1}0 & \cellcolor{col3}49 & \cellcolor{col1}0 & \cellcolor{col1}0 & \cellcolor{col1}0 & \cellcolor{col3}40 & \cellcolor{col1}0 & \cellcolor{col1}0 & \cellcolor{col1}0 & \cellcolor{col1}0 \\
\hline
\textbf{$g_8$} & \cellcolor{col2}12 & \cellcolor{col1}0 & \cellcolor{col1}5 & \cellcolor{col1}0 & \cellcolor{col1}0 & \cellcolor{col1}1 & \cellcolor{col1}0 & \cellcolor{col1}0 & \cellcolor{col1}0 & \cellcolor{col1}0 & \cellcolor{col1}0 \\
\hline
\textbf{$g_9$} & \cellcolor{col1}4 & \cellcolor{col1}0 & \cellcolor{col1}8 & \cellcolor{col1}0 & \cellcolor{col1}0 & \cellcolor{col1}0 & \cellcolor{col1}2 & \cellcolor{col1}0 & \cellcolor{col1}0 & \cellcolor{col1}0 & \cellcolor{col1}0 \\
\hline
\textbf{$g_{10}$} & \cellcolor{col2}12 & \cellcolor{col1}0 & \cellcolor{col1}2 & \cellcolor{col1}0 & \cellcolor{col1}0 & \cellcolor{col1}3 & \cellcolor{col1}0 & \cellcolor{col1}0 & \cellcolor{col1}0 & \cellcolor{col1}3 & \cellcolor{col1}1 \\
\hline
\textbf{$g_{11}$} & \cellcolor{col2}16 & \cellcolor{col1}1 & \cellcolor{col3}37 & \cellcolor{col1}0 & \cellcolor{col1}0 & \cellcolor{col1}4 & \cellcolor{col2}6 & \cellcolor{col1}0 & \cellcolor{col1}0 & \cellcolor{col1}3 & \cellcolor{col2}6 \\
\end{tabular}

}
\caption{$\mathcal{G}_x$ for \methodabbrv.} \label{fig:cm-encoder-based-c}
\end{subtable}
\end{minipage}%
\vspace{1.5em} 
\begin{minipage}{1.0\linewidth}
\centering
\begin{subtable}[t]{1\textwidth}
\centering
\setlength{\arrayrulewidth}{0.3mm}
\setlength{\tabcolsep}{5pt} 
\renewcommand{\arraystretch}{1.2}
\begin{tabular}{c|c|c}
$\mathcal{C}_x$ / $\mathcal{G}_x$   & \xmark & \cmark \\
\hline
\xmark & \edgebox{50} 539 & / \\
\hline
\cmark & \confounderbox{50} 423 & \idealbox{50}  3038 \\
\end{tabular}
\caption{Grounding Error Types.} \label{fig:cm-encoder-based-d}
\end{subtable}
\end{minipage}

\caption{\textbf{(a, b)} Confusion matrices on $\mathcal{C}_x$ for \stdabbrv and \methodabbrv. \textit{Neg.} and \textit{Pos.} indicate negative and positive class. \textbf{(c)} Confusion matrix on $\mathcal{G}_x$ for \methodabbrv. \textbf{(d)} Grounding error types matrix for \methodabbrv. \xmark and \cmark stand for wrong and correct prediction. Color coding refers to \Edge, \Ideal, and \Confounder cases.} \label{tab:cm-encoder-based}
\end{figure}

\paragraph{Quantitative Results.}
\Cref{tab:experiments:classification-merged} (right) reports classification performance.
Regarding binary classification task ($\mathcal{C}_x$), we observe a notable performance degradation in the \methodabbrv setting to the \stdabbrv one.
A possible motivation is that these models are fine-tuned on instructions that require $\mathcal{C}_x$ rather than $\mathcal{G}_x$ as an output~\cite{wang-etal-2022-super, blevins-etal-2023-prompting}.
Therefore, prompt examples compliant with \methodabbrv could fall into the out-of-domain boundary.
Moreover, LLMs significantly underperform to encoder-based models on all tasks, with a gap of around $\sim$15-20 percentage F$_1$ points.
This result is aligned with recent insights about LLMs on information extraction tasks, where fine-tuned encoder-based models are yet to be surpassed~\cite{zhou-etal-2024-universalner}.
We provide additional considerations regarding the performance gap in the following.

\paragraph{Qualitative Results.}


%

We inspect LLM-generated responses to better assess the performance gap observed in~\Cref{tab:experiments:classification-merged}.
A possible cause of performance degradation may be attributed to models not following reported instructions, leading to hallucinations or unrelated content~\cite{heo-etal-2024-llms-know-internally-follow}.
Interestingly, we do not observe such a behavior in either setting.
LLMs always follow prompt instructions correctly, except for a few cases ($<2.9\%$) in Phi under \stdabbrv where the model responds by selecting a guideline $g$ rather than answering as instructed.\footnote{We repeat \stdabbrv experiments without guidelines to assess their impact in Appendix~\ref{app:llm-no-guidelines}.}
Consequently, we analyze model predictions in \methodabbrv as in \S~\ref{sec:encoder-based} by computing the Grounding Error Types confusion matrix.
In EDOS, $60$-$71$\% of model predictions fall in the \Edge case, while $17$-$20$\% belong to \Confounder case.
In contrast, in ToS, \Edge cases increase up to $79$-$97$\% of the cases, while \Confounder cases are less than $1$\% in all categories.
These results further suggest that LLMs struggle to identify the right $\mathcal{G}_x$, underscoring their incapability in addressing the task as formulated in \methodabbrv.
Nonetheless, further studies concerning more advanced prompting settings are required to corroborate these observations, which we leave as future work.

\section{Discussion} \label{sec:discussion}

\paragraph{Alternative Annotation Methodology}
As an alternative to \methodabbrv, we could report \textit{both} $\mathcal{G}_x$ and $\mathcal{C}_x$ for each data sample.
This solution would allow for transparent adherence to annotation guidelines and fine-grained evaluation of annotations as in \methodabbrv.
However, this approach is exposed to an annotation shortcut where the annotator may first determine $\mathcal{C}_x$ and then define $\mathcal{G}_x$ accordingly, thus reversing the expected annotation process.
Furthermore, as in \stdabbrv, this alternative is tailored to a specific $\mathcal{C}$, implying that a change in $\mathcal{C}$ requires re-annotating collected data.
Lastly, reporting both $\mathcal{G}_x$ and $\mathcal{C}_x$ carries the annotation cost of two distinct annotations, which is not desired.

\paragraph{Iterative Guideline Refinement}
The prescriptive paradigm assumes that $\mathcal{G}$ is fixed during annotation.
However, in real-world scenarios, guidelines are subject to continuous refinement with human-in-the-loop helping in addressing edge cases as data is collected~\cite{van-der-stappen-etal-2021-guidelines-designing, guzman-monteza-2023-annotation-method}.
Consequently, data annotation is required at each guideline refinement iteration, leading to substantial efforts and costs.
Nonetheless, \methodabbrv transparency to guidelines via $\mathcal{G}_x$ might be preferable when moving from one iteration to another to limit re-annotation costs. 
In particular, information about $\mathcal{G}_x$ helps identify sample categories that change labeling across iterations, possibly derived from their inherent ambiguity, as it is the case in subjectivity detection~\cite{antici-etal-2024-corpus-sentence}.
Therefore, researchers can track how a task evolves through guideline refinement and similarly extract insights about annotation to how data cartography provides for models~\cite{swayamdipta-etal-2024-data-cartography}.

\paragraph{Guideline Availability}
A requirement of using \methodabbrv for annotation is that guidelines are always paired with published annotated data and are no longer optional material.
This has two important consequences.
First, it fosters reproducibility since researchers can easily understand how annotators have been instructed for data collection, thus, contributing to open research~\cite{pineau-etal-2021-reproducibility-ml}.
Second, it raises awareness on how and for what purpose data was collected, fostering responsible research and limiting wrong uses of the data.
For instance, several definitions of hate speech exist, each reflecting specific cultures~\cite{korre-etal-2024-untangling-hatespeech-definitions}.
Therefore, training models on several datasets that share the same $\mathcal{C}$, but have incompatible $\mathcal{G}$ should and must be avoided.

\paragraph{\methodabbrv in Other Annotation Paradigms}
\methodabbrv is compatible with the perspectivist approach~\cite{cabitza-et-al-2023-perspectivist-turn}, which does not foresee a convergence to one label: the label variations~\cite{plank-2022-problem} among annotators can be captured and entailed in the collected dataset and for model developments~\cite{davani-etal-2022-dealing, 10.1613/jair.1.12752}. 
In particular, capturing different annotators' perspectives by indicating which specific guideline was selected via \methodabbrv fosters a fine-grained comprehension of annotators' viewpoints and evaluation of collected annotations.
The applicability of \methodabbrv to other annotation paradigms not based on detailed guidelines, such as the descriptive one \cite{rottger-etal-2022-two}, is not straightforward and deserves a separate study, which we leave as future work.

\section{Conclusions} \label{sec:conclusion}
We presented \methodabbrv, a novel annotation methodology that reports annotation guidelines rather than class labels for each data sample.
The advantages \methodabbrv include annotation and model training adherence to guidelines, finer-grained analysis of annotated data, and support for multiple task annotations that rely on the same guidelines.
We showed that annotating with \methodabbrv has an increased annotation time overhead that is compensated by added insights about data quality and model evaluation. 
We hope \methodabbrv can foster researchers in developing high-quality benchmarks for discerning the decision-making process and the limitations of state-of-the-art models, an ever-recurring need in NLP~\cite{yadkori-etal-2024-believe}.
As future work, we aim to extend our analysis to other annotation paradigms like the descriptive one~\cite{rottger-etal-2022-two}, develop datasets from scratch using \methodabbrv, and, consequently, further explore model evaluation via $\mathcal{G}_x$ supervision.




\bibliography{references}
\bibliographystyle{acl_natbib}

\appendix

\section{Data Supplementary Details} 
\label{app:tos}

\Cref{tab:tos} reports statistics about ToS-100~\cite{ruggeri-et-al-2022-tos}.

\begin{table}[!h]
\centering
\small
\resizebox{\columnwidth}{!}{
\begin{tabular}{lrr}
    \toprule
    Clause type & No. Unfair clauses & $\vert \mathcal{G} \vert$\\
    \midrule
    Arbitration  (A)                        & 106 (0.50\%) & 8 \\
    Unilateral change (CH) 			        & 344 (1.68\%) & 7 \\
    Content removal (CR)	                & 216 (1.06\%) & 17 \\
    Limitation of liability (LTD)  			& 626 (3.07\%) & 18 \\
    Unilateral termination (TER)    	    & 420 (2.06\%) & 28 \\
    \bottomrule
\end{tabular}
}
\caption{ToS-100 statistics. We report the number and percentage of unfair clauses and the number of rationales for each unfairness category. \label{tab:tos}}
\end{table}

\section{Hardware and Implementation Details} 
\label{app:hardware}

We implemented all baselines and methods in PyTorch~\cite{paszke-etal-2019-pytorch} and PyTorch Lightning.\footnote{\url{https://github.com/Lightning-AI/pytorch-lightning}.}
All experiments were run on a private machine with an NVIDIA 3060Ti GPU with 8 GB dedicated VRAM.
LLMs undergo a 4-bit quantization process to reduce computational overhead and fit single-GPU training.
We will release all of the code and data to reproduce our experiments in an MIT-licensed public repository.

\section{Additional Text-based Architectures} \label{app:text-based-variants}

In the \textit{Text-based} setting, we consider two other architecture variants to evaluate the impact of different embedding strategies: \textit{Dot} and \textit{Concat}.
We denote the \textit{Text-based} model reported in the main paper as \textit{Entail} for distinction.
In \textit{Dot}, we compute the dot product of the embedding pair $(x, g)$ and feed it to the classification head. 
This is the original architecture of~\citet{pappas-henderson-2019}.
Similarly, \textit{Concat} replaces the dot product via embedding concatenation.
Results (\Cref{tab:experiments:text-based-variants}) show that \textit{Entail} outperforms other variants in all tasks.

\section{\stdabbrv Without Guidelines} \label{app:llm-no-guidelines}
We repeat \stdabbrv experiments by removing guidelines from the prompt to assess model sensitivity to guidelines and observe that LLM performance degrades significantly (\Cref{tab:experiments:llm-prompting-nog}).
This is a desired result since the involved benchmarks are domain-specific tasks.

\begin{table}[!h]
    \centering
    \small
    \resizebox{\columnwidth}{!}{
    \begin{tabular}{ll|rrr}
    \toprule
    \multicolumn{1}{c}{} & \multicolumn{1}{c|}{} & \multicolumn{1}{c}{\textit{Dot}} & \multicolumn{1}{c}{\textit{Concat}} & \multicolumn{1}{c}{\textit{Entail}} \\ \midrule
    \multirow{2}{*}{ToS$_{A}$}    & $\mathcal{C}_x$ & $67.07_{\pm 6.34}$ & $66.81_{\pm 6.78}$ & $\mathbf{80.14_{\pm 5.29}}$ \\
                                  & $\mathcal{G}_x$ & $18.31_{\pm 8.24}$ & $19.55_{\pm 7.44}$ & $\mathbf{37.49_{\pm 5.28}}$ \\
    \multirow{2}{*}{ToS$_{CH}$}   & $\mathcal{C}_x$ & $80.93_{\pm 4.14}$ & $86.13_{\pm 7.70}$ & $\mathbf{90.23_{\pm 1.84}}$ \\
                                  & $\mathcal{G}_x$ & $27.35_{\pm 2.94}$ & $18.02_{\pm 3.65}$ & $\mathbf{44.92_{\pm 9.93}}$ \\
    \multirow{2}{*}{ToS$_{CR}$}   & $\mathcal{C}_x$ & $67.85_{\pm 5.33}$ & $72.72_{\pm 5.51}$ & $\mathbf{78.14_{\pm 3.76}}$ \\
                                  & $\mathcal{G}_x$ & $18.21_{\pm 3.24}$ & $19.46_{\pm 3.72}$ & $\mathbf{27.21_{\pm 5.36}}$ \\
    \multirow{2}{*}{ToS$_{LTD}$}  & $\mathcal{C}_x$ & $72.82_{\pm 3.04}$ & $75.00_{\pm 1.96}$ & $\mathbf{80.39_{\pm 1.40}}$ \\
                                  & $\mathcal{G}_x$ & $17.12_{\pm 2.70}$ & $20.31_{\pm 3.23}$ & $\mathbf{21.49_{\pm 1.41}}$ \\
    \multirow{2}{*}{ToS$_{TER}$}  & $\mathcal{C}_x$ & $80.44_{\pm 4.19}$ & $77.64_{\pm 4.81}$ & $\mathbf{84.57_{\pm 4.94}}$ \\
                                  & $\mathcal{G}_x$ & $11.15_{\pm 1.50}$ & $10.38_{\pm 1.31}$ & $\mathbf{20.04_{\pm 3.08}}$ \\
    \multirow{2}{*}{EDOS}         & $\mathcal{C}_x$ & $71.10_{\pm 2.20}$ & $74.96_{\pm 2.20}$ & $\mathbf{79.87_{\pm 2.24}}$ \\
                                  & $\mathcal{G}_x$ & $\mathbf{24.70_{\pm 2.07}}$ & $18.19_{\pm 1.37}$ & $22.97_{\pm 3.14}$ \\ 
                                  \bottomrule
    \end{tabular}%
    }
    \caption{Text-based model performance on ToS and EDOS datasets. Best results are in \textbf{bold}.}
    \label{tab:experiments:text-based-variants}
\end{table}


\begin{table}[!h]
\centering
\scriptsize
\resizebox{1.0\linewidth}{!}{
\begin{tabular}{lrrr}
\toprule
\multicolumn{1}{l}{Dataset} & \multicolumn{1}{c}{Mistral} & \multicolumn{1}{c}{Llama} & \multicolumn{1}{c}{Phi} \\ \midrule
ToS$_{A}$    & $41.40$ ($\downarrow 15.67$) & $36.52$ ($\downarrow 26.20$) & $42.02$ ($\downarrow 22.08$) \\
ToS$_{CH}$   & $42.85$ ($\downarrow 19.58$) & $38.69$ ($\downarrow 22.11$) & $44.29$ ($\downarrow 17.90$) \\
ToS$_{CR}$   & $42.46$ ($\downarrow 13.30$) & $37.90$ ($\downarrow 18.74$) & $43.33$ ($\downarrow 16.63$) \\
ToS$_{LTD}$  & $45.74$ ($\downarrow 14.97$) & $40.84$ ($\downarrow 14.95$) & $44.85$ ($\downarrow 13.15$) \\
ToS$_{TER}$  & $44.66$ ($\downarrow 18.40$) & $40.12$ ($\downarrow 14.74$) & $46.36$ ($\downarrow 11.45$) \\
EDOS         & $40.90$ ($\downarrow 15.57$) & $59.72$ ($\downarrow 08.21$) & $60.88$ ($\downarrow 02.59$) \\ \bottomrule
\end{tabular}
}
\caption{LLMs performance on ToS and EDOS datasets when no guideline information is provided in the prompt. Brackets indicate performance degradation with respect to~\Cref{tab:experiments:classification-merged}.}
\label{tab:experiments:llm-prompting-nog}
\end{table}

\end{document}